\documentclass[letterpaper]{article} 
\usepackage[submission]{aaai25}  
\usepackage{times}  
\usepackage{helvet}  
\usepackage{courier}  
\usepackage[hyphens]{url}  
\usepackage{graphicx} 
\usepackage{natbib}  
\usepackage{caption} 
\usepackage{algorithm}
\usepackage{algorithmic}
\usepackage{amssymb}
\usepackage{colortbl}
\usepackage{bm}
\usepackage{amsmath}
\usepackage{multirow} 
\usepackage{adjustbox}
\usepackage{makecell}
\usepackage{stackengine}
\usepackage{subcaption}
\usepackage{graphicx}
\usepackage{booktabs}
\newcommand\xrowht[2][0]{\addstackgap[.5\dimexpr#2\relax]{\vphantom{#1}}}
\usepackage{xcolor}
\usepackage{graphicx}
\usepackage{newfloat}
\usepackage{listings}
\DeclareCaptionStyle{ruled}{labelfont=normalfont,labelsep=colon,strut=off} 
\floatstyle{ruled}
\newfloat{listing}{tb}{lst}{}
\floatname{listing}{Listing}
\makeatletter
\def\showauthors@on{T}
\makeatother

\title{Learning from Different Samples: A Source-free Framework for Semi-supervised Domain Adaptation
}
\author {
    Xinyang Huang\textsuperscript{\rm 1},
    Chuang Zhu$^\dag$\textsuperscript{\rm 1},
    Bowen Zhang\textsuperscript{\rm 1},
    Shanghang Zhang\textsuperscript{\rm 2}
}
\affiliations {
    \textsuperscript{\rm 1}Beijing University of Posts and Telecommunications\\
    \textsuperscript{\rm 2}Peking University\\
    hsinyanghuang7@gmail.com
}

\begin{document}

\maketitle

\begin{abstract}
Semi-supervised domain adaptation (SSDA) has been widely studied due to its ability to utilize a few labeled target data to improve the generalization ability of the model.
However, existing methods only consider designing certain strategies for target samples to adapt, ignoring the exploration of customized learning for different target samples.
When the model encounters complex target distribution, existing methods will perform limited due to the inability to clearly and comprehensively learn the knowledge of multiple types of target samples.
To fill this gap, this paper focuses on designing a framework to use different strategies for comprehensively mining different target samples.
We propose a novel source-free framework (SOUF) to achieve semi-supervised fine-tuning of the source pre-trained model on the target domain.
Different from existing SSDA methods, SOUF decouples SSDA from the perspectives of different target samples, specifically designing robust learning techniques for unlabeled, reliably labeled, and noisy pseudo-labeled target samples.
For unlabeled target samples, probability-based weighted contrastive learning (PWC) helps the model learn more discriminative feature representations.
To mine the latent knowledge of labeled target samples, reliability-based mixup contrastive learning (RMC) learns complex knowledge from the constructed reliable sample set.
Finally, predictive regularization learning (PR) further mitigates the misleading effect of noisy pseudo-labeled samples on the model.
Extensive experiments on benchmark datasets demonstrate the superiority of our framework over state-of-the-art methods.
Our code will be available after acceptance.
\end{abstract}

\section{Introduction}
\label{sec:intro}
Deep neural networks (DNNs) have achieved great success in many computer vision tasks \cite{krizhevsky2012imagenet,krizhevsky2017imagenet,simonyan2014very}. 
However, these networks need to provide rich labels during training \cite{rawat2017deep,alzubaidi2021review}.
Therefore, domain adaptation (DA) is proposed to alleviate this problem and help the model migrate from a source domain with rich labels to a target domain with different distributions of no or few labels \cite{ganin2015unsupervised,pan2010domain}.
Depending on the number of target labels, DA can be divided into unsupervised domain adaptation (UDA) and semi-supervised domain adaptation (SSDA).
This paper focuses on SSDA, which can utilize a few target labels to expand semantic information and learn target knowledge, thereby achieving domain alignment.

\begin{figure}[t]
	\centering
\includegraphics[width=1\linewidth]{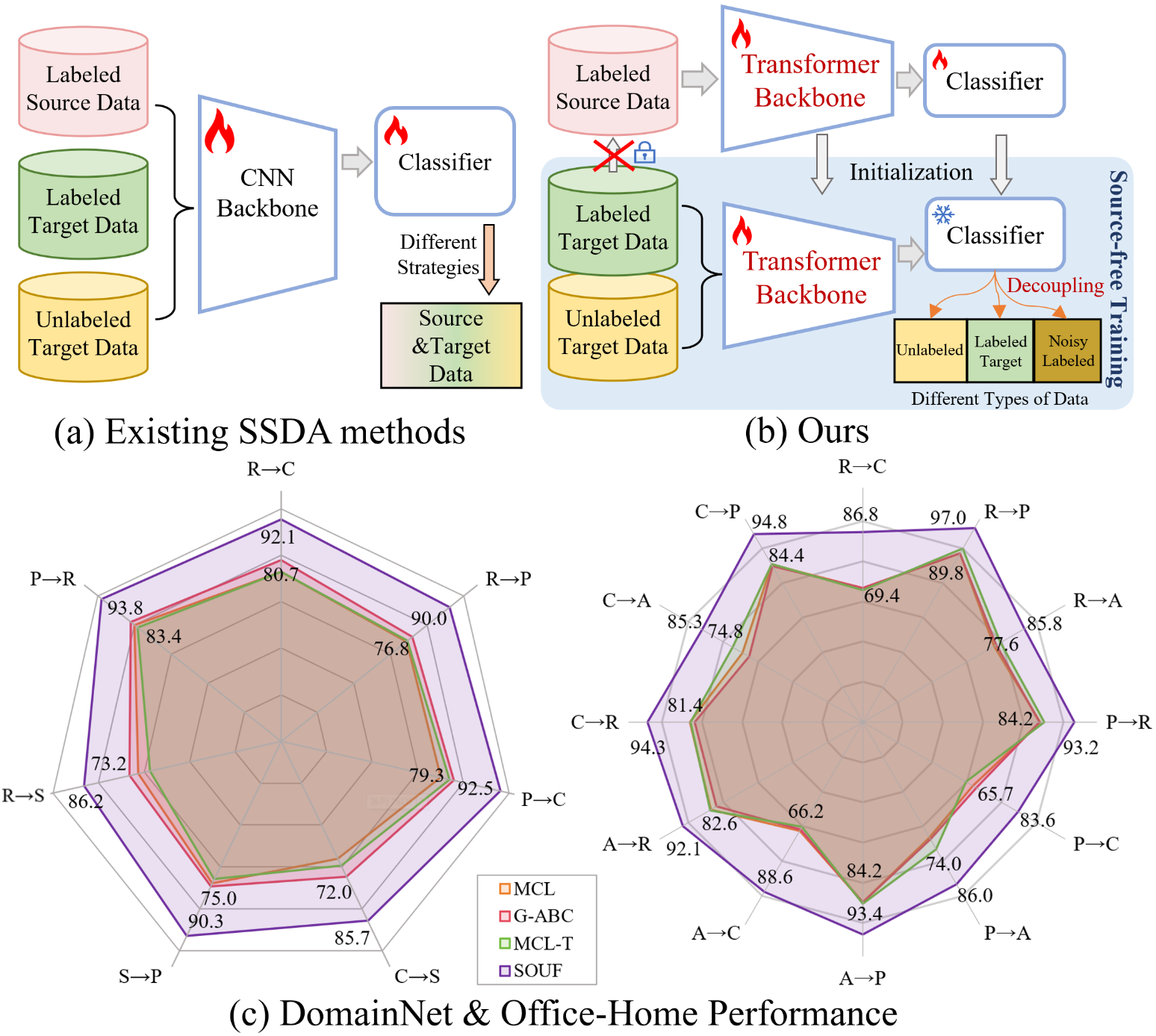}
	\caption{Differences between the proposed method and existing methods.
 Figures (a) and (b) show that our method focuses on fine-tuning in a semi-supervised target domain without the source data.
 Based on this learning framework, SOUF performs customized learning for different types of target samples.
Figure (c) shows that our method achieves SoTA with a large advantage on the DomainNet and Office-Home datasets.
}
	\label{intro}
\end{figure}

Due to its practical advantages, SSDA has attracted more attention and has been widely studied \cite{saito2019semi,li2021cross,li2021ecacl,yan2022multi,yu2023semi,huang2023semi,li2023adaptive,li2024inter,he2024enhancing}.
However, existing methods focus on the study of target sample learning strategies and ignore the importance of customized learning for different types of target samples.
Most existing methods \cite{yan2022multi,huang2023semi,yu2023semi} only design a series of learning strategies for unlabeled samples, but when there is \textit{knowledge bias} in the distribution of unlabeled target samples (such as noisy pseudo-label samples), the learned knowledge is unreliable.
At the same time, existing methods \cite{yan2022multi,li2023adaptive} ignore the potential deep knowledge of labeled target samples, which is difficult to learn with simple supervised learning.
Therefore, it is necessary to design a framework to use different strategies for the comprehensive mining of different target samples.
To fill this gap, this paper proposes a novel source-free framework called \textbf{SOUF}, which decouples the SSDA from the perspective of learning different samples.
Decoupling the target domain according to different samples can help the model better understand the target domain by learning different sub-tasks.
Compared with existing SSDA methods, our framework can more fully utilize the powerful representation and transfer capabilities of the model, thereby better adapting to semi-supervised target domains.

As shown in Figure \ref{intro}, unlike the training paradigm of most existing SSDA methods, this paper considers a source-free scenario \cite{liang2020we}, $i.e.$, fine-tuning the source-pretrained model in the target domain.
Our SOUF is proposed to decouple the semi-supervised target domain from the perspective of learning from unlabeled, reliable labeled, and noisy pseudo-labeled target samples.

Specifically, we propose probability-based weighted contrastive learning (PWC) for unlabeled target samples to help the model learn more discriminative representations.
The adaptive weights assign lower weights to low-confidence samples to reduce the impact of erroneous semantic information, which can better help the model learn unlabeled target knowledge \cite{zhang2023towards}.
In the semi-supervised fine-tuning stage, it is also essential to further learn the latent knowledge of labeled samples.
We combine labeled samples with high-confidence unlabeled samples to construct a new set of reliable labeled samples.
We propose reliability-based mixup contrastive learning (RMC) to mix transformer patches from the constructed reliable labeled sample set and learn complex target representations, which can capture the commonalities and differences between different samples \cite{zhang2017mixup,chen2022transmix,zhu2023patch}.
Finally, even if we learn the feature knowledge of labeled and unlabeled samples through sufficient contrastive learning, the learned knowledge will still be biased when there is much noise in target pseudo-labeled samples \cite{song2019prestopping, liu2020early}.
We leverage predictions of pseudo-labeled samples to constrain the probabilistic output from the predictive regularization learning (PR) perspective to improve the performance of the model when facing complex target data.

Our contributions can be summarized as follows:
1) We propose a novel source-free framework for SSDA scenarios to decouple and customize learning of different types of target samples.
2) We design a series of novel learning methods for unlabeled, reliably labeled, and noisy pseudo-labeled target samples.
3) Our framework is one of the first attempts to solve SSDA using a source-free transformer-based framework.
4) Compared with existing SSDA methods, our method achieves state-of-the-art performance and does not need to access any source data during the semi-supervised adaptation stage.

\section{Related Work}
\label{related work}
\subsection{Unsupervised Domain Adaptation}
To solve the problem that supervised learning requires extensive manual annotation, unsupervised domain adaptation (UDA) aims to transfer knowledge from a fully labeled source domain to an unlabeled target domain.
The basic idea is to use the maximum mean difference (MMD)\cite{gretton2012kernel}, which achieves migration from the source domain to the target domain by minimizing the feature distribution.
DANN \cite{ganin2016domain} and JAN \cite{long2017deep} propose using the MMD criterion to learn transmission networks through cross-region alignment of multiple region-specific layers.
Meanwhile, many recent works \cite{long2018conditional,hoffman2018cycada,xie2018learning,shen2018wasserstein,ge2023unsupervised} perform domain alignment from the perspective of adversarial learning can be effectively exploited to the target domain.
Compared with CNN-based methods, due to the good generalization ability of the transformer \cite{vaswani2017attention}, many works \cite{yang2023tvt,xu2021cdtrans,sun2022safe,zhu2023patch} have also explored the application of the transformer in UDA.
However, due to the ability to utilize additional target domain information, SSDA methods usually perform better than UDA methods in the target domain.

\subsection{Semi-supervised Domain Adaptation}
Semi-supervised domain adaptation (SSDA) aims to use a small number of labeled target samples to help the model better adapt to the target domain.
Existing works can be simply divided into cross-domain alignment, adversarial training, and semi-supervised learning methods.
In terms of cross-domain alignment, related works \cite{li2021ecacl,singh2021improving,yang2021deep,he2024enhancing} integrate various complementary domain alignment techniques.
G-ABC \cite{li2023adaptive} further achieves semantic alignment by forcing the transfer from labeled source and target data to unlabeled target samples.
Utilizing the idea of adversarial training, related methods \cite{saito2019semi,qin2021contradictory,jiang2020bidirectional,kim2020attract,li2021cross,qin2022semi,ma2022context} solve the SSDA by minimizing the entropy between the prototype and adjacent unlabeled target domain samples, thereby achieving adversarial training.
Through semi-supervised learning methods, MCL \cite{yan2022multi} and ProML \cite{huang2023semi} further help the model understand target domains that lack a large number of labels through consistency regularization.
However, existing methods ignore the importance of customized learning for different types of target samples. 
This paper decouples SSDA and proposes a learning framework called SOUF to fully learn the target domain from the perspective of learning different samples.

\begin{figure*}[!ht]
	\centering
\includegraphics[width=0.8\linewidth]{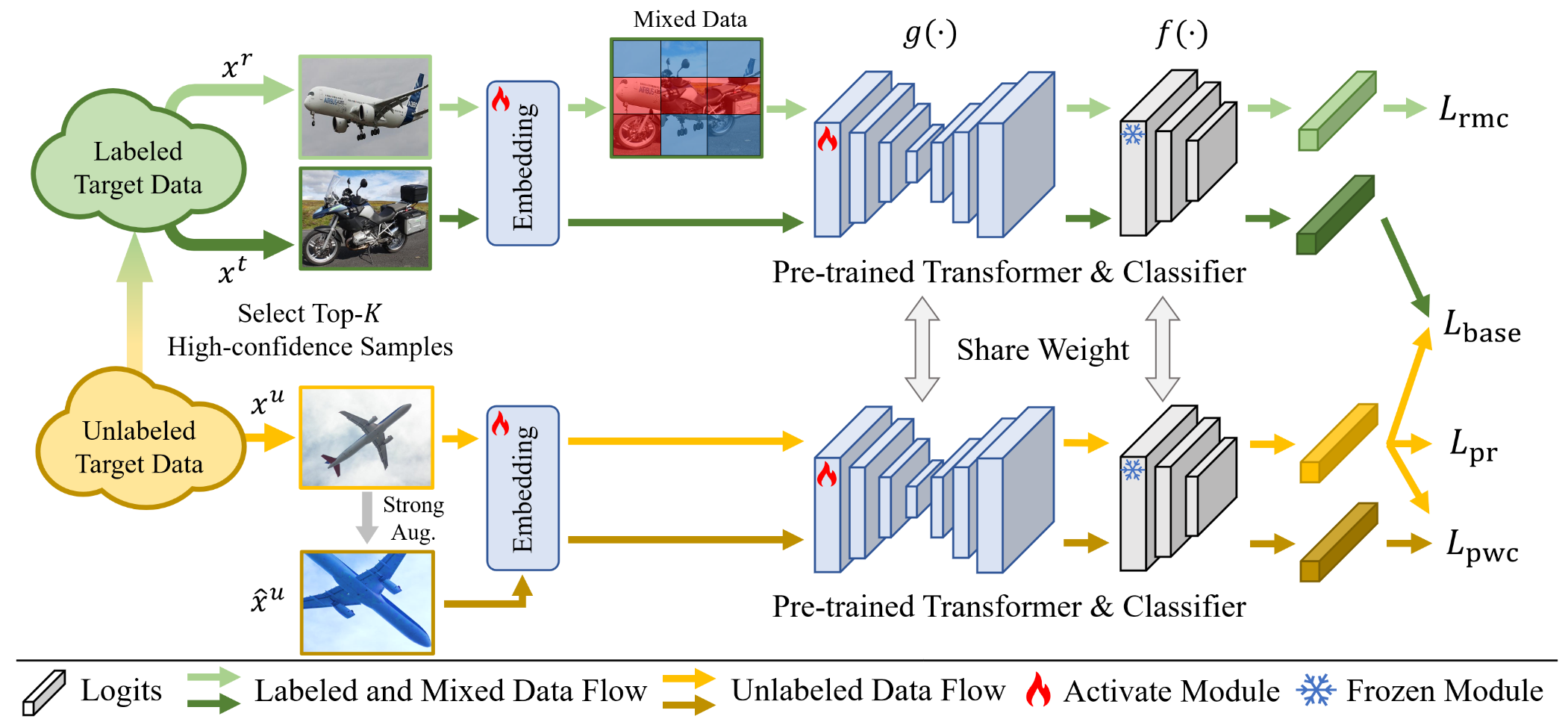}
	\caption{Illustration of our source-free framework. 
 For unlabeled samples, probability-weighted contrastive learning (PWC) adaptively learns discriminative features, enhancing credible probability outputs. 
 To further exploit labeled samples, reliability-based mixup contrastive learning (RMC) mixes patches from reliable samples, aiding in complex representation learning. 
 Predictive regularization (PR) minimizes the impact of erroneous semantic information from noisy labels.
 \vspace{-3mm}}
\label{overview}
\end{figure*}

\section{Methodology}
\label{methodology}
\subsection{Preliminaries}
In the SSDA, the source dataset ${\mathcal{D}_s = \{x^s_i, y^s_i\}^{N_s}_{i=1}}$ contains fully labeled data, ${\mathcal{D}_t = \{x^t_i, y^t_i\}^{N_t}_{i=1}}$ contains a small amount of labeled data of the target domain, where $N_s$ and $N_t$ are the source domain and target domain dataset size respectively.
$x_i^s$ and $x_i^t$ represent the labeled source and target image data and $y^s_i$ and $y^t_i$ represent the corresponding labels.
There is also an unlabeled target image set ${\mathcal{D}_u = \{x^u_i\}^{N_u}_{i=1}}$ for adaptation in the target domain, which contains unlabeled target image data and $N_u\gg N_t$.
Our framework comprises a transformer-based feature extractor $g(\cdot)$ and a linear classifier $f(\cdot)$, as shown in Figure \ref{overview}.
Following \cite{huang2023semi}, we generate the strong augment view for each unlabeled target sample $x_{i}^u$, represented as $\hat{x}_{i}^{u}$.
The target samples are then fed to the same feature extractor $g(\cdot)$ and classifier $f(\cdot)$ to obtain the probabilistic predictions $p_{i}^{u}$, $\hat{p} _{i}^{u}$.
This paper uses the cross-entropy loss to train feature extractor $g(\cdot)$ and classifier $f(\cdot)$ on the source domain.
Following works \cite{liang2020we,ma2022context}, we freeze $f(\cdot)$ and train $g(\cdot)$ in the target domain.
Based on this, we will introduce the proposed learning framework and how the training program can achieve further target learning.

\subsection{Probability-based Weighted Contrastive Learning}
After pre-training in the source domain, we consider fine-tuning the feature extractor in the semi-supervised target domain. 
Therefore, an effective method is needed to help the model learn the target knowledge representation in a large number of unlabeled target samples.
In recent years, contrastive learning \cite{oord2018representation,grill2020bootstrap,he2020momentum} has been proven to be a reliable representation learning method, which can help models better understand data and learn helpful knowledge representations by constraining feature representations between samples.
However, feature-based contrastive learning cannot represent the actual target distribution represented by unlabeled target features, which will harm the generalization ability of the classifier in the target domain.

To address these challenges, we consider proposing probability-based weighted contrastive learning (PWC) to help the model better adapt to the target domain.
Specifically, we consider the following loss:
\begin{equation}
\scalebox{0.85}{$
\begin{aligned}
L_{\mathrm{pwc}}&=-\sum_{i=1}^{2N_u}\sum_{k=1}^{2N_u}w_{ik}\log\frac{\exp(p_i^u\cdot p_k^{u+}/\tau)}{\sum_{j=1}^{2N_u}\mathbb{I}(j\neq i)\mathrm{exp}(p_i^u\cdot p_j^u/\tau)},
\end{aligned}
$}
\end{equation}
where $p^{u+}_k$ represents the predicted probability of the positive target sample $k$, $\mathbb{I}(\cdot)$ represents the indicator function, $\tau$ represents the temperature coefficient, and the adaptive weight $w_{ik}$ is defined as follows:
\begin{equation}
\label{adaweight}
\scalebox{0.9}{$
\left.w_{ik}=\left\{
\begin{array}{ll}
1 & \text{if }k=i, \\
p_i^u\cdot p_k^u & \text{if $i$ and $k$ belong to the same category}, \\
0 & \text{otherwise}.
\end{array}
\right.\right.
$}
\end{equation}

Compared with instance-level contrastive learning, PWC learns category-level target relationships and can better combine the good feature representation extracted by the model.
When considering the similarity between two samples, only when the predicted probability of two similar samples of the same category is the maximum value 1, their probability product $p_i^u\cdot p_k^u$ is 1.
Therefore, the probability-based contrastive learning constraint model makes low-entropy judgments for similar target samples, thereby learning more discriminative feature representations. 
In addition, the adaptive weight $w$ in Eq.~\ref{adaweight}
helps the model learn more accurate sample relationships in the target domain by reducing the weight of low-confidence sample pairs and the possible error in learning target sample relationships.

\subsection{Reliability-based Mixup Contrastive Learning}
In the target adaptation stage, it is necessary to leverage the model to learn the latent knowledge of labeled samples \cite{berthelot2019mixmatch}, which is ignored by existing SSDA methods.
Not limited to labeled samples, we consider combining labeled and high-confidence unlabeled samples to build a new set of reliable labeled samples to learn complex target knowledge.

\noindent\textbf{Constructing reliable labeled samples.}
We mix image patches from new sets of reliable labeled samples to construct mixed samples to learn complex knowledge representations. 
The knowledge learned by mixing samples from the reliably labeled sample set is more reliable than directly mixing unlabeled samples.
Because learning complex target knowledge relies on reliable sample label information \cite{yao2018deep,wang2022reliable,albert2021relab,chen2023refining}, the pseudo-labels of low-confidence unlabeled samples may contain noise, and directly mixing them will cause the model to learn incorrect target knowledge.
Specifically, the new reliable sample set is constructed as follows:
\begin{equation}
\scalebox{0.9}{$
\begin{aligned}
\mathcal{D}_{reliable}
=\mathcal{D}_{t}\cup \mathcal{D}_{high-conf}
=\{x^r_i, y^r_i\}^{N_t+C\times K}_{i=1}
=\{x^r_i, y^r_i\}^{N_r}_{i=1},
\end{aligned}
$}
\end{equation}
where $\mathcal{D}_{t}$ represents the labeled sample set, $C$ represents the total number of categories, $\mathcal{D}_{high-conf} = \{x^h_i, y^h_i\}^{C\times K}_{i=1}$ represents the top-$K$ unlabeled samples with the smallest entropy in each category $c$, and $y^h_i$ represents the pseudo-label of $x^h_i$ output by the model.

For samples from the new reliable sample set $\mathcal{D}_{reliable}$, we mix image patches from two different samples:
\begin{equation}
\begin{aligned}
&\mathcal{X}_{kn}=\lambda_{n}\cdot x_{in}^{r}+(1-\lambda_{n})\cdot x_{jn}^{r},
\end{aligned}
\end{equation}
where $x_{in}^{r}$ represents the $n$-th patch of the $i$-th sample from the reliable label sample set $\mathcal{D}_{reliable}$, $\mathcal{X}_{kn}$ represents the $n$-th patch of the $k$-th mixed sample, and $\lambda_{n}$ is sampled from a learnable beta distribution $\mathrm{Beta}(\beta,\gamma)$.

\noindent\textbf{Mixup cross-entropy learning.}
For mixed samples $\mathcal{X}_k$, we first consider calculating its mixup cross-entropy loss \cite{zhang2017mixup}:
\begin{equation}
\scalebox{0.9}{$
\begin{aligned}
L_{\mathrm{mix-ce}}=\sum_{k=1}^{N_r}\left(\hat{\lambda}_k\cdot y^r_{i}\log(\mathcal{P}_k)+(1-\hat{\lambda}_k)\cdot y^r_{j}\log(\mathcal{P}_k)\right),
\end{aligned}
$}
\end{equation}
where $\mathcal{P}_k$ represents the probability output of $\mathcal{X}_k$, and $y^r_{i}$ and $y^r_{j}$ represent the labels of components $x^r_{i}$ and $x^r_{j}$ for the mixed samples $\mathcal{X}_k$, and $\hat{\lambda}_k$ represents the adaptive coefficient after rescaling the attention score in the transformer:
\begin{equation}
\begin{aligned}
\hat{\lambda}_k=\frac{\sum_{n=1}^m\lambda_na_{in}}{\sum_{n=1}^m\lambda_na_{in}+\sum_{n=1}^m(1-\lambda_n)a_{jn}},\\
\end{aligned}
\end{equation}
where $a_{in}$ represents the average of the sum of attention scores for each layer corresponding to the $n$-th patch, and $m$ is the number of the patches.

\noindent\textbf{Mixup contrastive learning.}
Furthermore, we consider a novel mixup contrastive learning based on the PWC proposed in the previous section to learn complex target relationships.
The specific equation is as follows:
\begin{equation}
\scalebox{0.8}{$
\begin{aligned}
L_{\mathrm{mix-con}}=
&-\sum_{k=1}^{2N_r}\hat{\lambda}_k\sum_{i=1}^{2N_r}w_{ki}\log\frac{\exp(\mathcal{P}_k\cdot p_i^{r}/\tau)}{\sum_{i=1}^{2N_r}\mathbb{I}(i\neq k)\mathrm{exp}(\mathcal{P}_k\cdot p_i^r/\tau)}\\
&-\sum_{k=1}^{2N_r}(1-\hat{\lambda}_k)\sum_{j=1}^{2N_r}w_{kj}\log\frac{\exp(\mathcal{P}_k\cdot p_j^{r}/\tau)}{\sum_{j=1}^{2N_r}\mathbb{I}(j\neq k)\mathrm{exp}(\mathcal{P}_k\cdot p_j^r/\tau)},
\end{aligned}
$}
\end{equation}
where $p^r_{i}$ and $p^r_{j}$ represent the probability outputs of components $x^r_{i}$ and $x^r_{j}$.
Combining the mixup cross-entropy loss and mixup contrastive learning loss, we consider the complete reliability-based mixup contrastive loss (RMC) as follows:
\begin{equation}
\begin{aligned}
L_{\mathrm{rmc}}=L_{\mathrm{mix-ce}}+L_{\mathrm{mix-con}}.
\end{aligned}
\end{equation}

Since original contrastive learning relies heavily on discriminative semantic knowledge, there is not enough opportunity to learn from negative examples once the model can distinguish between positive and negative examples.
Even if applying certain data augmentation techniques can alleviate this problem, as proposed by works \cite{chen2020simple,wang2022contrastive,zhang2022rethinking}, original contrastive learning is not robust to different data augmentation combinations.
Through the proposed RMC, the model can customizedly learn the latent knowledge of labeled samples in the target domain, thereby better and comprehensively adapting to the target domain.

\subsection{Predictive Regularization Learning}
Even if the framework sufficiently considers comprehensive contrastive learning, the learning of the model will still be biased when there is more noise in the target pseudo-label. 
However, existing SSDA methods omit this effect.
The early training phenomenon \cite{song2019prestopping,liu2020early,bai2021understanding} confirms that classifiers can predict noisy pseudo-labeled samples with high accuracy in the early adaptation stage before overfitting noisy data.
Therefore, we propose predictive regularization learning (PR) for the SSDA scenario for the first time to reduce the misleading impact of noisy pseudo-labeled samples on the model.
The following equation gives the specific form:
\begin{equation}
\label{elr}
L_\mathrm{pr}=\frac1{N_{u}}\sum_{i=1}^{N_{u}}\log(1-\hat{y}^{u\ell\top}_{i} p_i^{u\ell}),
\end{equation}
where $p_i^{u\ell}$ is the target probability output at $\ell$-th epoch, $\hat{y}^{u\ell}_i = \alpha\cdot \hat{y}^{u(\ell-1)}_i+(1-\alpha)\cdot p_i^{u\ell}$ is the moving average prediction and $\alpha=0.7$ is the hyper-parameter. 
Compared with the existing SSDA method, PR can help the model better combine the proposed contrastive learning technique and minimize the impact of noisy labels.

\subsection{Overall Training Objective}
For labeled and unlabeled target data, we employ the cross-entropy objective for their labels and pseudo-labels, respectively, and combine them with mutual information loss \cite{liang2020we}.
We denote it as $L_\mathrm{base}$ and leave its details to \cite{liang2020we} due to the page limit.
Finally, the overall loss function can be shown as follows:
\begin{equation}
\label{allloss}
L_\mathrm{all}=L_\mathrm{base}+\lambda_\mathrm{pwc}L_\mathrm{pwc}+\lambda_\mathrm{rmc}L_\mathrm{rmc}+\lambda_\mathrm{pr}L_\mathrm{pr},
\end{equation}
where $\lambda_\mathrm{pwc}$, $\lambda_\mathrm{rmc}$ and $\lambda_\mathrm{pr}$ are scalar hyper-parameters of the loss weights.
By completely decoupling the target domain for semi-supervised adaptation, the framework can better leverage the complementary advantages of different learning objectives, thereby fully improving the generalization ability of the model.
The overall target adaptation algorithm is described in the supplementary material.

\begin{table*}[t]
\centering
\renewcommand\arraystretch{0.98}
\begin{adjustbox}{width=\textwidth}
\scalebox{0.8}{
\begin{tabular}{cc|c|cccccccccccccc|cc}
\toprule
\multirow{2}{*}{SF}&\multirow{2}{*}{Trans.}&\multirow{2}{*}{Method} & \multicolumn{2}{c}{R$\rightarrow$C} & \multicolumn{2}{c}{R$\rightarrow$P} & \multicolumn{2}{c}{P$\rightarrow$C} & \multicolumn{2}{c}{C$\rightarrow$S} & \multicolumn{2}{c}{S$\rightarrow$P} & \multicolumn{2}{c}{R$\rightarrow$S} & \multicolumn{2}{c}{P$\rightarrow$R} & \multicolumn{2}{c}{Mean}\\
~ & ~ & ~ & 1-shot & 3-shot & 1-shot & 3-shot & 1-shot & 3-shot & 1-shot & 3-shot& 1-shot & 3-shot & 1-shot & 3-shot& 1-shot & 3-shot & 1-shot & 3-shot\\
\midrule
$\times$&$\times$&S+T&55.6&60.0&60.6&62.2&56.8&59.4&50.8&55.0&56.0&59.5&46.3&50.1&71.8&73.9&56.9&60.0\\
$\times$&$\times$&DANN&58.2&59.8&61.4&62.8&56.3&59.6&52.8&55.4&57.4&59.9&52.2&54.9&70.3&72.2&58.4&60.7\\
$\times$&$\times$&MME&70.0&72.2&67.7&69.7&69.0&71.7&56.3&61.8&64.8&66.8&61.0&61.9&76.1&78.5&66.4&68.9\\
$\times$&$\times$&CDAC&77.4&79.6&74.2&75.1&75.5&79.3&67.6&69.9&71.0&73.4&69.2&72.5&80.4&81.9&73.6&76.0\\
$\times$&$\times$&CLDA&76.1&77.7&75.1&75.7&71.0&76.4&63.7&69.7&70.2&73.7&67.1&71.1&80.1&82.9&71.9&75.3\\
$\times$&$\times$&ECACL&75.3&79.0&74.1&77.3&75.3&79.4&65.0&70.6&72.1&74.6&68.1&71.6&79.7&82.4&72.8&76.4\\
$\times$&$\times$&ASDA&77.0&79.4&75.4&76.7&75.5&78.3&66.5&70.2&72.1&74.2&70.9&72.1&79.7&82.3&73.9&76.2\\
$\times$&$\times$&MCL&77.4&79.4&74.6&76.3&75.5&78.8&66.4&70.9&74.0&74.7&70.7&72.3&82.0&83.3&74.4&76.5\\
$\times$&$\times$&ProML&78.5&80.2&75.4&76.5&77.8&78.9&70.2&72.0&74.1&75.4&72.4&73.5&84.0&84.8&76.1&77.4\\
$\times$&$\times$&SLA&79.8&81.6&75.6&76.0&77.4&80.3&68.1&71.3&71.7&73.5&71.7&73.5&80.4&82.5&75.0&76.9\\
$\times$&$\times$&IDMNE&79.6&80.8&76.0&76.9&79.4&80.3&71.7&72.2&75.4&75.4&73.5&73.9&82.1&82.8&76.8&77.5\\
$\times$&$\times$&G-ABC& 80.7&82.1&76.8&76.7&79.3&81.6&72.0&73.7&75.0&76.3& 73.2&74.3& 83.4 &83.9& 77.5 & 78.2 \\
$\times$&$\times$&TriCT& 86.5 &89.1 &85.3 &86.6 &80.4 &86.3 &71.0 &79.9 &80.3 &84.5 &78.9 &82.1 &82.2& 90.1& 80.7 &85.5\\
\checkmark&$\times$&DEEM& 79.7 &80.5 & 78.1 &79.0 & 77.0 &77.5& 71.9 &74.9 & 77.7 &80.0 & 76.7 &75.9 & 85.4 &88.5 & 78.1 & 79.5 \\
\checkmark & $\times$ & SOUF-C (Ours) & 87.0 & 88.5 & 85.2 & 87.0 & 87.7 & 88.2 & 81.8 & 82.7 & 85.5 & 86.8 & 81.0 & 82.8 & 89.8 & 90.7 & 85.4 & 86.6\\
\hline
$\times$&\checkmark&MCL-T$^\dag$& 77.3 & 79.8 & 75.0 & 76.5 & 78.0 & 80.0 & 68.6 & 71.9 & 72.7 & 75.6 & 67.3 & 70.5 & 81.0 & 83.3 & 74.3 & 76.8 \\ 
\checkmark&\checkmark&SHOT-T$^\dag$& 74.3 & 76.6 & 78.8 & 82.4 & 75.4 & 78.8 & 70.5 & 71.0 & 78.6 & 81.9 & 69.9 & 72.1 & 86.5 & 87.5 & 76.3 & 78.6 \\
\checkmark&\checkmark&DEEM-T$^\dag$& 82.9 &83.8 & 81.3 &82.1 & 80.3 &80.5 & 75.8 &77.9 & 80.9 &83.5 & 80.2 &80.4 & 88.7 &89.9 & 81.4 & 82.5 \\
\rowcolor{gray!25} \checkmark&\checkmark&SOUF (Ours)&\textbf{92.1}&\textbf{93.1}&\textbf{90.0}&\textbf{92.3}&\textbf{92.5}&\textbf{92.8}&\textbf{85.7}&\textbf{86.2}&\textbf{90.3}&\textbf{91.2}&\textbf{86.2}&\textbf{87.7}&\textbf{93.8}&\textbf{94.0}&\textbf{90.0} \textcolor{red}{(+8.6)}&\textbf{91.0} \textcolor{red}{(+5.5)}\\
\bottomrule
\end{tabular}}
\end{adjustbox}
\caption{Accuracy (\%) on \textit{\textit{DomainNet}} under the settings of 1-shot and 3-shot; $SF$ represents the source-free training scenario; $Trans.$ represents the transformer-based method and $\dag$ represents our reproduced results.
\vspace{-1mm}}
\label{domainnet1}
\end{table*}

\begin{table*}[ht]
\centering
\renewcommand\arraystretch{0.95}
\scalebox{0.8}{
\begin{tabular}{cc|c|cccccccccccc|c}
\toprule
SF & Trans. & Method   & R$\rightarrow$C   & R$\rightarrow$P  & R$\rightarrow$A    & P$\rightarrow$R    & P$\rightarrow$C  & P$\rightarrow$A      & A$\rightarrow$P  & A$\rightarrow$C  & A$\rightarrow$R  & C$\rightarrow$R   & C$\rightarrow$A   & C$\rightarrow$P  & Mean  \\ 
\midrule
$\times$&$\times$&S+T & 52.1 & 78.6 & 66.2 & 74.4 & 48.3 & 57.2 & 69.8 & 50.9 & 73.8 & 70.0 & 56.3 & 68.1 & 63.8  \\
$\times$&$\times$&DANN& 53.1 & 74.8 & 64.5 & 68.4 & 51.9 & 55.7 & 67.9 & 52.3 & 73.9 & 69.2 & 54.1 & 66.8 & 62.7  \\ 
$\times$&$\times$&MME& 61.9 & 82.8 & 71.2 & 79.2 & 57.4 & 64.7 & 75.5 & 59.6 & 77.8 & 74.8 & 65.7 & 74.5 & 70.4  \\
$\times$&$\times$&APE& 60.7 & 81.6 & 72.5 & 78.6 & 58.3 & 63.6 & 76.1 & 53.9 & 75.2 & 72.3 & 63.6 & 69.8 & 68.9  \\ 
$\times$&$\times$&CDAC& 61.9 & 83.1 & 72.7 & 80.0 & 59.3 & 64.6 & 75.9 & 61.2 & 78.5 & 75.3 & 64.5 & 75.1 & 71.0  \\ 
$\times$&$\times$&SLA& 66.1 &84.6 &72.7 &80.5 &61.8 &67.3 &78.0 &63.0 &79.2 &77.0 &66.9 &77.6 & 72.9\\
$\times$&$\times$&MCL& 67.0 & 85.5 & 73.8 & 81.3 & 61.1 & 68.0 & 79.5 & 64.4 & 81.2 & 78.4 & 68.5 & 79.3 & 74.0  \\ 
$\times$&$\times$&ProML & 67.5 &86.1 &73.7 &81.9& 61.4 &69.3 &79.7 &64.5 &81.7 &79.0 &69.1 &80.5 &74.6 \\
\checkmark&$\times$&DEEM$^\dag$& 62.0 &82.5 & 71.6 &77.1 & 60.8 &65.8& 74.5 &63.9 & 75.0 &76.5 & 62.0 &76.8 & 70.7 \\
\checkmark&$\times$&SOUF-C~(Ours)& 79.9 & 89.5 & 80.6 & 84.2 & 77.9 & 79.7 & 85.7 & 82.5 & 84.1 & 85.4 & 84.7 & 86.6& 83.4\\
\hline
$\times$&\checkmark&MCL-T$^\dag$& 59.9 & 85.0 & 75.5 & 82.4 & 57.8 & 72.1 & 79.3 & 58.3 & 79.7 & 78.5 & 69.3 & 78.9 & 73.1 \\ 
\checkmark&\checkmark&SHOT-T$^\dag$& 65.8 & 77.5 & 68.8 & 75.8 & 64.8 & 68.9 & 76.7 & 63.3 & 75.9 & 78.0 & 74.6 & 80.3 & 72.6 \\ 
\checkmark&\checkmark&DEEM-T$^\dag$& 69.9 & 81.2 & 72.6 & 79.5 & 69.2 & 73.1 & 80.4 & 68.3 & 79.6 & 80.6 & 78.4 & 83.7 & 76.4 \\ 
\rowcolor{gray!25} \checkmark&\checkmark&SOUF~(Ours)& \textbf{84.4}& \textbf{94.5}& \textbf{85.1}& \textbf{89.2}& \textbf{82.4} & \textbf{84.2} & \textbf{90.2} & \textbf{87.5} & \textbf{89.1}& \textbf{89.9} & \textbf{84.9}  & \textbf{91.1}&\textbf{87.7} \textcolor{red}{(+11.3)}\\
\bottomrule
\end{tabular}}
\caption{Accuracy (\%) on \textit{Office-Home} under the setting of 1-shot; $SF$ represents the source-free training scenario; $Trans.$ represents the transformer-based method and $\dag$ represents our reproduced results.
\vspace{-3mm}}
\label{office-home1}
\end{table*}

\begin{table*}[ht]
\centering
\renewcommand\arraystretch{0.95}
\scalebox{0.8}{
\begin{tabular}{cc|c|cccccccccccc|c}
\toprule
SF & Trans. & Method   & R$\rightarrow$C   & R$\rightarrow$P  & R$\rightarrow$A    & P$\rightarrow$R    & P$\rightarrow$C  & P$\rightarrow$A      & A$\rightarrow$P  & A$\rightarrow$C  & A$\rightarrow$R  & C$\rightarrow$R   & C$\rightarrow$A   & C$\rightarrow$P  & Mean  \\ 
\midrule
$\times$&$\times$&S+T & 55.7 & 80.8 & 67.8 & 73.1 & 53.8 & 63.5 & 73.1 & 54.0 & 74.2 & 68.3 & 57.6 & 72.3 & 66.2 \\ 
$\times$&$\times$&DANN&57.3 &75.5 &65.2 &69.2 &51.8 &56.6& 68.3 &54.7 &73.8 &67.1 &55.1 &67.5 &63.5\\
$\times$&$\times$&MME& 64.6 &85.5& 71.3 &80.1 &64.6 &65.5 &79.0 &63.6 &79.7 &76.6 &67.2 &79.3 &73.1\\
$\times$&$\times$&APE& 66.4 &86.2 &73.4 &82.0 &65.2 &66.1 &81.1 &63.9 &80.2 &76.8 &66.6 &79.9 &74.0\\
$\times$&$\times$&CDAC& 67.8 & 85.6 & 72.2 & 81.9 & 67.0 & 67.5 & 80.3 & 65.9 & 80.6 & 80.2 & 67.4 & 81.4 & 74.2  \\ 
$\times$&$\times$&SLA& 70.1 & 87.1 & 73.9 & 82.5 & 69.3 & 70.1 & 82.6 & 67.3 & 81.4 & 80.1 & 69.2 & 82.1 & 76.3  \\ 
$\times$&$\times$&MCL& 70.1 & 88.1 & 75.3 & 83.0 & 68.0 & 69.9 & 83.9 & 67.5 & 82.4 & 81.6 & 71.4 & 84.3 & 77.1  \\ 
$\times$&$\times$&ProML& 71.0 &88.6 &75.8 &83.8 &68.9 &72.5 &83.9& 67.8& 82.2& 82.3& 72.1 &84.1& 77.8\\
$\times$&$\times$&IDMNE& 71.7 &88.1& 75.2& 82.7& 67.6& 69.0& 82.4& 66.4& 79.3&79.5& 69.1& 83.1& 76.2\\
$\times$&$\times$&G-ABC& 70.0& 88.1& 76.0& 82.8& 69.3& 70.5& 83.8& 67.2& 80.4& 80.2& 69.2& 83.9 & 77.2 \\
$\times$&$\times$&TriCT& 81.9 &94.1 &\textbf{86.3} &92.3 &78.7 &83.4 &91.1 &76.9 &91.3 &91.8 &80.5 &92.5 &86.7\\
\checkmark&$\times$&DEEM$^\dag$& 67.2 &83.0 & 67.1 &78.3 & 69.3 &66.3& 79.7 &65.9 & 78.3 &77.6 & 65.1 &80.8 & 73.2 \\
\checkmark&$\times$&SOUF-C (Ours)& 84.3   & 94.0   & 82.6  & 89.9  & 81.4   & 83.1  & 89.8   & 86.0    & 89.3   & 91.2  & 82.1    & 92.1   & 87.1\\
\hline
$\times$&\checkmark&MCL-T$^\dag$& 69.4 & 89.8 & 77.6 & 84.2 & 65.7 & 74.0 & 84.2 & 66.2 & 82.6 & 81.4 & 74.8 & 84.4 & 77.9 \\ 
\checkmark&\checkmark&SHOT-T$^\dag$& 71.5 & 86.3 & 75.7 & 80.5 & 69.2 & 71.9 & 80.6 & 69.1 & 68.8 & 79.4 & 74.5 & 81.6 & 76.6 \\
\checkmark&\checkmark&DEEM-T$^\dag$& 75.3 & 89.7 & 79.3 & 84.4 & 73.1 & 76.0 & 84.3 & 73.5 & 82.6 & 83.2 & 78.1 & 85.3 & 80.4 \\ 
\rowcolor{gray!25}
\checkmark&\checkmark&SOUF (Ours)& \textbf{86.8}   & \textbf{97.0}   & 85.8  & \textbf{93.2}  & \textbf{83.6}   & \textbf{86.0}  & \textbf{93.4}   & \textbf{88.6}    & \textbf{92.1}   & \textbf{94.3}  & \textbf{85.3}    & \textbf{94.8}   & \textbf{90.1} \textcolor{red}{(+3.4)}\\
\bottomrule
\end{tabular}}
\caption{Accuracy (\%) on \textit{Office-Home} under the setting of 3-shot with DeiT-S \cite{touvron2021training} backbone which has the similar parameter size as ResNet-34 \cite{he2016deep} used by existing CNN-based methods; $SF$ represents the source-free training scenario; $Trans.$ represents the transformer-based method and $\dag$ represents our reproduced results.
\vspace{-3mm}}
\label{office-home3}
\end{table*}

\begin{table}[t]
\centering
\renewcommand\arraystretch{0.95}
\scalebox{0.85}{
\begin{tabular}{cc|c|cc|c}
    \hline
    SF & Trans. & Method & D$\rightarrow$A & W$\rightarrow$A & Mean \\ 
    \hline
    $\times$ & $\times$ &  S+T & 62.4 & 61.2  & 61.8   \\ 
    $\times$ & $\times$ &  DANN & 70.4 & 74.6  & 72.5   \\ 
    $\times$ & $\times$ &  MME & 73.6 & 77.6  & 75.6   \\
    $\times$ & $\times$ &  DECOTA & 74.2 & 78.3 & 76.3   \\
    \hline
    \checkmark & $\times$ &  DEEM$^\dag$ & 78.9 & 80.5 & 79.7 \\
    \rowcolor{gray!25}
    \checkmark & \checkmark &  SOUF (Ours) & \textbf{82.6} & \textbf{82.9} & \textbf{82.8} \\
    \hline
\end{tabular}}
\caption{Accuracy (\%) on \textit{Office-31} under the settings of 3-shot with DeiT-S \cite{touvron2021training} backbone which has the similar parameter size as ResNet-34 \cite{he2016deep} used by existing CNN-based methods.
\vspace{-3mm}}
\label{office-31}
\end{table}

\begin{center}
\begin{table}[t]
\centering
\renewcommand\arraystretch{1}
\tabcolsep=6.0pt
\scalebox{0.85}{
\begin{tabular}{c|ccc|cc|c}
    \hline
    Number & $L_\mathrm{pwc}$ & $L_\mathrm{rmc}$ & $L_\mathrm{pr}$ & P$\rightarrow$C & P$\rightarrow$R & Mean  \\ 
    \hline 
        1 &  &  &  & 79.9 & 86.5 &83.2 \\ 
        2 & \checkmark &  &  & 88.1 & 88.9 &88.5 \\ 
        3 &  & \checkmark &  & 83.6 & 86.9 &84.3 \\ 
        4 &  &  & \checkmark & 85.8 & 87.8 &86.3 \\ 
        5 & \checkmark & \checkmark &  & 89.4 & 90.8 & 90.1 \\ 
        6 & \checkmark &  & \checkmark & 90.9 & 91.6 & 91.3 \\ 
        7 &  & \checkmark & \checkmark & 86.4 & 88.2 & 87.3 \\ 
        8 & \checkmark & \checkmark & \checkmark & \textbf{92.5} & \textbf{93.8} & \textbf{93.2}\\
    \hline
    \end{tabular}}
    \caption{Accuracy (\%) of ablation study on \textit{DomainNet} under the setting of 1-shot.
    \vspace{-4mm}}
\label{absMainComponent}
\end{table}
\end{center}

\section{Experiment}
\label{experiment}

In this section, we conduct experiments to answer the following questions: 1) Does SOUF outperform existing SSDA methods? 2) Are the proposed novel components effective? 3) Does SOUF perform better than other transformer-based DA methods or CNN-based SSDA methods that replace the backbone with transformers?

\subsection{Experimental Settings}
The experiment mainly consists of three datasets. 
\textit{DomainNet} \cite{peng2019moment} is a large-scale domain adaptation benchmark dataset.
We selected four domains: Real (R), Clipart (C), Painting (P), and Sketch (S).
\textit{Office-Home} \cite{venkateswara2017deep} is a medium-sized SSDA benchmark dataset.
It includes four domains: Art (A), Clipart (C), Product (P), and Real (R).
\textit{Office-31} \cite{saenko2010adapting} is a small domain adaptation benchmark dataset.
We choose DeiT-S \cite{touvron2021training} as the extraction backbone of the transformer because its parameter size is similar to ResNet-34 \cite{he2016deep} used by most existing CNN-based methods.
The learning rate of the feature extractor classifier is set to 0.001 and 0.01.
The temperature coefficient $\tau$ in PWC is 0.15.
The number of high-confidence samples $K$ is 2.
The loss weights $\lambda_\mathrm{pwc}$, $\lambda_\mathrm{rmc}$ and $\lambda_\mathrm{pr}$ are specified as 0.1, 0.1 and 3 respectively.
We adopt the widely used Randaugmnt \cite{cubuk2020randaugment} as the strong data augmentation strategy.
For more details, please refer to our code and supplementary material.

\subsection{Comparison with State-of-the-Arts}
We compare the performance of SOUF with previous state-of-the-art SSDA methods, including S+T, DANN~\cite{ganin2016domain}, MME~\cite{saito2019semi}, APE~\cite{kim2020attract}, DECOTA~\cite{yang2021deep}, ECACL~\cite{li2021ecacl}, ASDA~\cite{qin2022semi}, MCL~\cite{yan2022multi}, SLA~\cite{yu2023semi}, CLDA~\cite{singh2021clda}, CDAC~\cite{li2021cross}, ProML~\cite{huang2023semi}, DEEM~\cite{ma2022context}, 
TriCT~\cite{ngo2023improved}, IDMNE~\cite{li2024inter}, G-ABC~\cite{li2023adaptive}, and SHOT~\cite{liang2020we}.
S+T refers to the method of training a model by supervising only labeled samples from two domains.
SHOT is a source-free UDA method, and we add target cross-entropy loss to adapt it to SSDA.
To further eliminate the unfairness of experiments using different backbone networks and further prove that our framework improves the ability of the model, we reproduce the results of MCL, SHOT, and DEEM on DeiT-S~\cite{touvron2021training} and denote them as ``-T$^\dag$''.
Due to the robustness of our method, we also apply PWC and PR learning techniques to the same CNN structure as existing methods, denoted as ``SOUF-C''.

\noindent\textbf{Effectiveness of the proposed SOUF.}
Table \ref{domainnet1}, \ref{office-home1}, \ref{office-home3} and \ref{office-31} show the quantitative comparison results of our proposed method with existing methods.
As can be seen from the results, our method far outperforms most previous methods in most scenarios without source data.
At the same time, the comparison results of SOUF-C with existing methods also show that our method can be well adapted to different backbone networks.
It is worth noting that, attributed to the strong potential of the model exploited by our framework, even if the backbone of the existing SSDA method is simply replaced with the transformer (such as MCL-T and DEEM-T), the performance is \textbf{still far} behind our SOUF.
The results show that our method performs well in different domains, further demonstrating the robustness of our method and the importance of learning from different samples.

\subsection{Ablation Study}
\textbf{Each main component in SOUF.}
We performed an ablation study of the main components of SOUF in the 1-shot and 3-shot setups of DomainNet P$\rightarrow$C and P$\rightarrow$R, as shown in Table \ref{absMainComponent}.
Rows 2-4 show that each component can produce significant improvements.
Rows 5-7 show that each combination can still improve performance, demonstrating the versatility of the proposed modules.
Meanwhile, the PWC and PR modules can bring more significant improvements to the model than the RMC module.
This is because, in the PWC and PR modules, the model has already learned good feature representations for most of the samples in the target domain, while the number of labeled target samples is relatively small.
The best performance is obtained when all components of the model are activated.
\vspace{-7mm}

\begin{center}
\begin{table}[t]
\centering
\renewcommand\arraystretch{1}
\tabcolsep=4.0pt
\scalebox{0.8}{
\begin{tabular}{ccc|ccc}
    \hline\xrowht{20pt}
\makecell[c]{Class-wise} & 
        \makecell[c]{Probability \\ Contrast} & \makecell[c]{Adaptive \\ Weight} &  \makecell[c]{DomainNet \\ C$\rightarrow$S} & \makecell[c]{Office-Home \\ R$\rightarrow$C} & \makecell[c]{Mean} \\ \hline
          & &  & 77.9 & 76.2&77.1 \\ 
         \checkmark & &  & 80.2 & 79.8&80.0 \\ 
        \checkmark & \checkmark & & 83.6 & 83.4 &83.5 \\ 
        \checkmark &\checkmark & \checkmark & \textbf{85.7} & \textbf{84.4}& \textbf{85.1}\\ \hline
    \end{tabular}}
    \caption{Accuracy (\%) of ablation study for different components in PWC with 1-shot setting.
    \vspace{-3mm}}
\label{ProbabilityContrastandAdaptiveWeight}
\end{table}
\end{center}

\begin{figure*}[t]
    \centering
	  \subfloat[]{\includegraphics[width=0.3\linewidth]{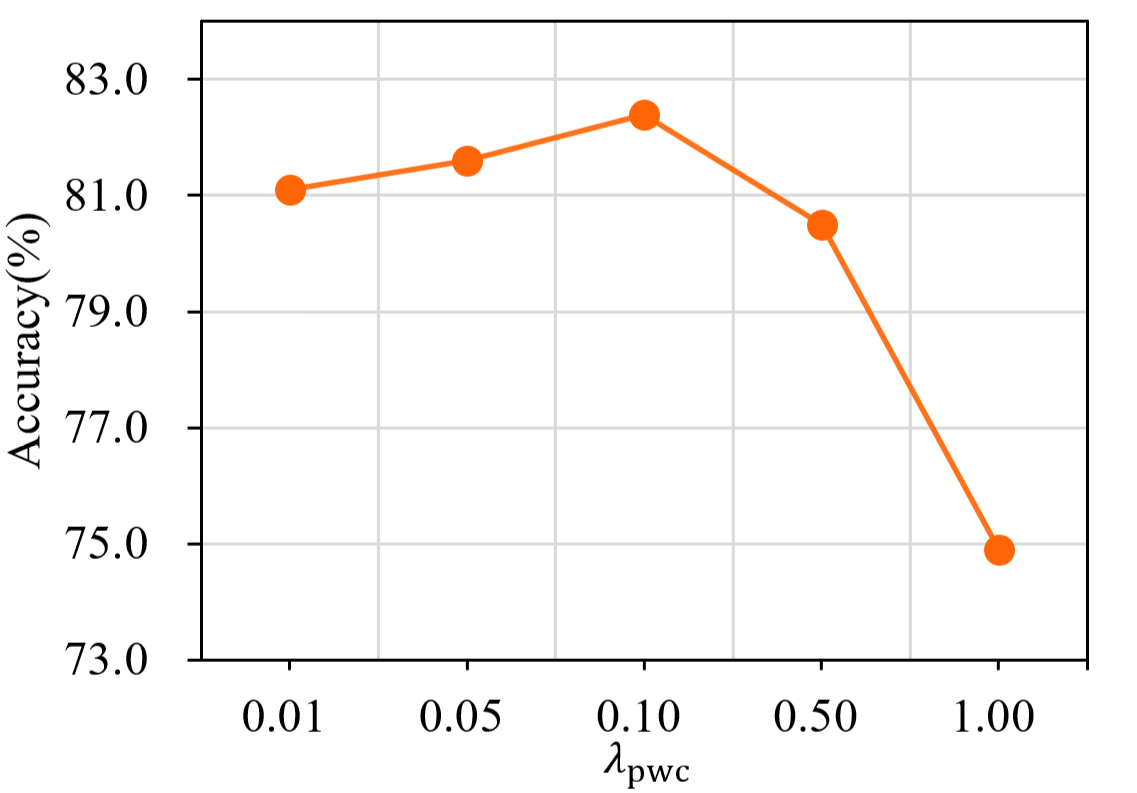}}
    \label{1a}\hfill
	  \subfloat[]{\includegraphics[width=0.3\linewidth]{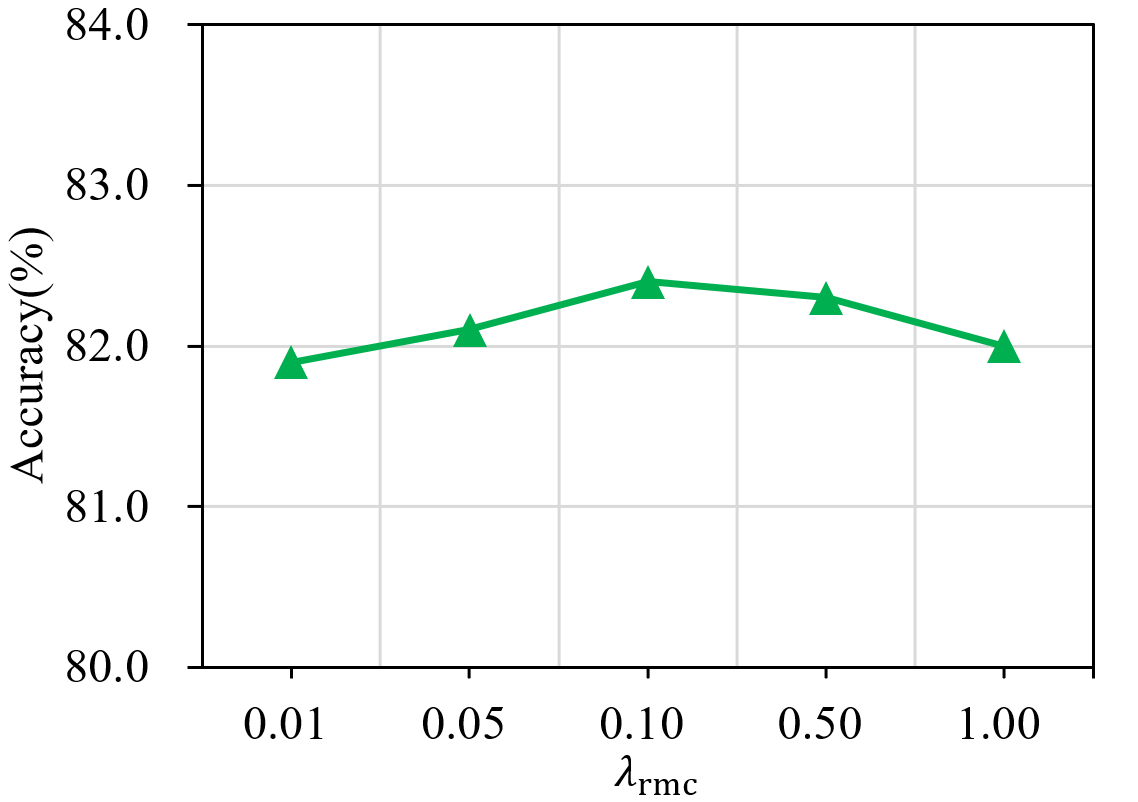}}
    \label{1b}\hfill
	  \subfloat[]{\includegraphics[width=0.3\linewidth]{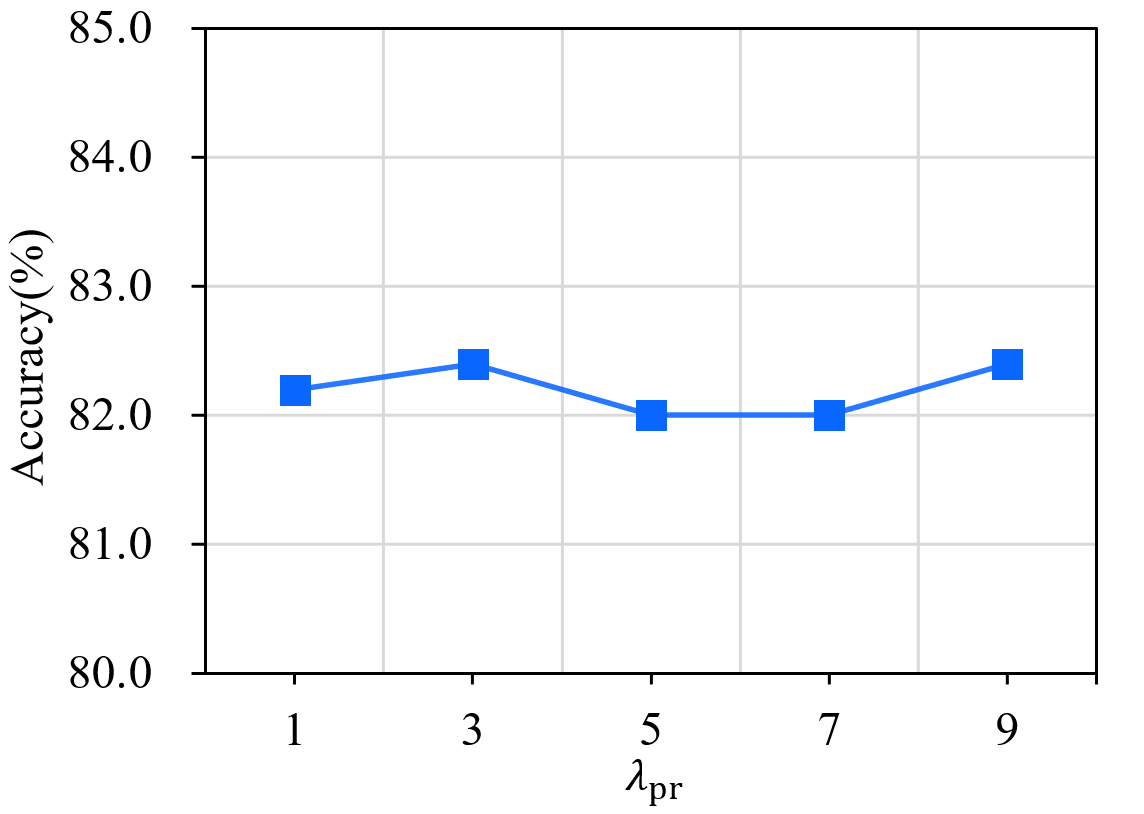}}
	  \caption{The effect of different loss balance parameters $\lambda_\mathrm{pwc}$, $\lambda_\mathrm{rmc}$, and $\lambda_\mathrm{pr}$ on the model classification accuracy in the \textit{Office-Home} P$\rightarrow$C scenario under the 1-shot setting.
   \vspace{-4mm}}
	  \label{senmainloss} 
\end{figure*}

\noindent\textbf{Each main component in PWC.}
We investigated the specific techniques mentioned in PWC to further demonstrate the effectiveness of PWC in our framework, as shown in Table \ref{ProbabilityContrastandAdaptiveWeight}.
It is worth noting that when no factors are considered, the model degenerates into InfoNCE loss \cite{chen2020simple}.
When considering contrastive learning between categories, the performance is improved because the model can learn the target knowledge between categories.
When learning discriminative features from the probability space is considered, the model performance improves significantly as the model is forced to output more confident predictions and can be combined with the knowledge learned by the classifier to allow the feature extractor to discover more compact clusters of target representations.
When considering the addition of adaptive weights, the model can adaptively learn relevant target representations of the same category with different confidence levels to achieve the best performance.

\noindent\textbf{Each main component in RMC.}
We show the performance impact of using only unlabeled samples, labeled samples, and building new reliable sample sets on RMC in Table \ref{reset1}.
RMC can further explore more target feature knowledge by building a new reliable sample set, thereby performing better in the target domain.
At the same time, we explore the impact of mixup contrastive learning and different mixup methods on performance in Table \ref{reset2}.
It can be seen that mixup contrastive learning can help the model learn more useful knowledge.
From the comparison of the last two rows, it can be verified that the performance of patch-level mixup is better than that of image-level mixup.
Patch-level mixup can re-weight according to the importance of each image patch, instead of linearly interpolating at the same ratio as image-level input.
It can leverage knowledge in transformer patches that are difficult for CNN to capture and better complement other proposed learning techniques.
This is also an \textbf{important} reason why SOUF can be specifically adapted to transformer-based backbone.

\begin{table}[t]
\centering
\renewcommand\arraystretch{0.95}
\tabcolsep=4.0pt
\scalebox{0.79}{
\begin{tabular}{ccc|ccc}
\toprule
Unlabeled & Labeled & \makecell[c]{Reliable (Ours)} & C$\rightarrow$A & C$\rightarrow$R & Mean \\
\midrule
\checkmark &  &  & 82.3 & 86.8 & 84.6 \\
 & \checkmark &  & 84.0 & 88.3 & 86.2 \\
 &  & \checkmark & \textbf{84.9} & \textbf{89.9} & \textbf{87.4} \\
\bottomrule
\end{tabular}}
\caption{Accuracy (\%) of ablation study with different samples for RMC on \textit{Office-Home} with 1-shot setting.}
\vspace{-3mm}
\label{reset1}
\end{table}

\begin{table}[t]
\centering
\renewcommand\arraystretch{0.95}
\tabcolsep=4.0pt
\scalebox{0.79}{
\begin{tabular}{ccc|ccc}
\toprule
\makecell[c]{Mixup contrast \\ learning} & \makecell[c]{Image-level \\ mixup} & \makecell[c]{Patch-level \\ mixup} & C$\rightarrow$A & C$\rightarrow$R & Mean \\
\midrule
 & & \checkmark & 83.9 & 88.6 & 86.3 \\
\checkmark & \checkmark&  & 82.4 & 87.8 & 85.1 \\
\checkmark & & \checkmark & \textbf{84.9} & \textbf{89.9} & \textbf{87.4} \\
\bottomrule
\end{tabular}}
\caption{Accuracy (\%) of ablation study with different mixup methods for RMC on \textit{Office-Home} with 1-shot task.
\vspace{-5mm}}
\label{reset2}
\end{table}

\subsection{Further Analysis}
\textbf{Comparison with transformer-based models in other DA scenarios.}
We note that there are works \cite{xu2021cdtrans} exploring the application of transformers in other DA scenarios (such as CDTrans \cite{xu2021cdtrans} in UDA).
We add the target domain supervision loss to CDTrans to apply it to the SSDA scenario and make a fair comparison with our method, as shown in Table \ref{clip2}.
Due to the lack of sufficient knowledge of the semi-supervised target domain, simply applying existing transformer-based methods in other DA scenarios to SSDA shows \textbf{limited} results compared to our SOUF, which further proves the innovation and effectiveness of our process for the SSDA.

\begin{table}[t]
\centering
\renewcommand\arraystretch{0.95}
\scalebox{0.79}{
\begin{tabular}{c|c|ccc}
\toprule
Method & Scenario &\makecell[c]{DomainNet \\ C$\rightarrow$S} & \makecell[c]{Office-Home \\ R$\rightarrow$C} &Mean\\
\midrule
CDTrans & UDA & 82.2 & 81.7 & 82.0 \\
SOUF (ours) & SSDA & \textbf{85.7} &\textbf{84.4} &\textbf{85.1} \\
\bottomrule
\end{tabular}}
\caption{Compare with transformer-based methods of other DA scenarios in SSDA.
\vspace{-5mm}}
\label{clip2}
\end{table}

\noindent\textbf{Sensitivity of $\lambda_\mathrm{pwc}$, $\lambda_\mathrm{rmc}$ and $\lambda_\mathrm{pr}$.}
In Eq. \ref{allloss}, we present the overall learning objectives of our SOUF framework which encompasses various components critical to the adaptation process.
To further explore the factors affecting the performance of our framework, we show the impact of the loss balance parameters $\lambda_\mathrm{pwc}$, $\lambda_\mathrm{rmc}$ and $\lambda_\mathrm{pr}$ on the classification accuracy under the Office-Home P$\rightarrow$C scenario in Figure \ref{senmainloss}.
Specifically, it can be observed that when $\lambda_\mathrm{pwc}=0.1$, $\lambda_\mathrm{rmc}=0.1$, and $\lambda_\mathrm{pr}=3$, the trained model achieves the highest performance in target image classification.
More analyses are in the supplementary material.

\section{Conclusion}
\label{conclusion}
In this paper, we present SOUF, a novel source-free framework for SSDA. 
Unlike existing methods, SOUF is proposed to decouple SSDA into distinct learning tasks for unlabeled, reliably labeled, and noisy pseudo-labeled target samples. 
For unlabeled samples, probability-weighted contrastive learning (PWC) enhances the model's ability to learn discriminative features. 
Reliability-based mixup contrastive learning (RMC) further enriches the model's understanding by mixing patches from a reliable sample set, capturing complex target knowledge. 
Predictive regularization learning (PR) mitigates the impact of noisy pseudo-labels by aligning current predictions with earlier outputs. 
Extensive experiments validate that SOUF significantly outperforms state-of-the-art SSDA methods.
\bibliography{aaai25}

\begin{thebibliography}{62}
\providecommand{\natexlab}[1]{#1}

\bibitem[{Albert et~al.(2021)Albert, Ortego, Arazo, O'Connor, and McGuinness}]{albert2021relab}
Albert, P.; Ortego, D.; Arazo, E.; O'Connor, N.; and McGuinness, K. 2021.
\newblock Relab: Reliable label bootstrapping for semi-supervised learning.
\newblock In \emph{2021 International Joint Conference on Neural Networks (IJCNN)}, 1--8. IEEE.

\bibitem[{Alzubaidi et~al.(2021)Alzubaidi, Zhang, Humaidi, Al-Dujaili, Duan, Al-Shamma, Santamar{\'\i}a, Fadhel, Al-Amidie, and Farhan}]{alzubaidi2021review}
Alzubaidi, L.; Zhang, J.; Humaidi, A.~J.; Al-Dujaili, A.; Duan, Y.; Al-Shamma, O.; Santamar{\'\i}a, J.; Fadhel, M.~A.; Al-Amidie, M.; and Farhan, L. 2021.
\newblock Review of deep learning: Concepts, CNN architectures, challenges, applications, future directions.
\newblock \emph{Journal of big Data}, 8: 1--74.

\bibitem[{Bai et~al.(2021)Bai, Yang, Han, Yang, Li, Mao, Niu, and Liu}]{bai2021understanding}
Bai, Y.; Yang, E.; Han, B.; Yang, Y.; Li, J.; Mao, Y.; Niu, G.; and Liu, T. 2021.
\newblock Understanding and improving early stopping for learning with noisy labels.
\newblock \emph{Advances in Neural Information Processing Systems}, 34: 24392--24403.

\bibitem[{Berthelot et~al.(2019)Berthelot, Carlini, Goodfellow, Papernot, Oliver, and Raffel}]{berthelot2019mixmatch}
Berthelot, D.; Carlini, N.; Goodfellow, I.; Papernot, N.; Oliver, A.; and Raffel, C.~A. 2019.
\newblock Mixmatch: A holistic approach to semi-supervised learning.
\newblock \emph{Advances in neural information processing systems}, 32.

\bibitem[{Chen et~al.(2022)Chen, Sun, He, Torr, Yuille, and Bai}]{chen2022transmix}
Chen, J.-N.; Sun, S.; He, J.; Torr, P.~H.; Yuille, A.; and Bai, S. 2022.
\newblock Transmix: Attend to mix for vision transformers.
\newblock In \emph{Proceedings of the IEEE/CVF Conference on Computer Vision and Pattern Recognition}, 12135--12144.

\bibitem[{Chen et~al.(2020)Chen, Kornblith, Norouzi, and Hinton}]{chen2020simple}
Chen, T.; Kornblith, S.; Norouzi, M.; and Hinton, G. 2020.
\newblock A simple framework for contrastive learning of visual representations.
\newblock In \emph{International conference on machine learning}, 1597--1607. PMLR.

\bibitem[{Chen et~al.(2023)Chen, Liu, Wang, Wang, Liu, and Wang}]{chen2023refining}
Chen, Y.; Liu, M.; Wang, X.; Wang, F.; Liu, A.-A.; and Wang, Y. 2023.
\newblock Refining Noisy Labels with Label Reliability Perception for Person Re-identification.
\newblock \emph{IEEE Transactions on Multimedia}.

\bibitem[{Cubuk et~al.(2020)Cubuk, Zoph, Shlens, and Le}]{cubuk2020randaugment}
Cubuk, E.~D.; Zoph, B.; Shlens, J.; and Le, Q.~V. 2020.
\newblock Randaugment: Practical automated data augmentation with a reduced search space.
\newblock In \emph{Proceedings of the IEEE/CVF conference on computer vision and pattern recognition workshops}, 702--703.

\bibitem[{Ganin and Lempitsky(2015)}]{ganin2015unsupervised}
Ganin, Y.; and Lempitsky, V. 2015.
\newblock Unsupervised domain adaptation by backpropagation.
\newblock In \emph{International conference on machine learning}, 1180--1189. PMLR.

\bibitem[{Ganin et~al.(2016)Ganin, Ustinova, Ajakan, Germain, Larochelle, Laviolette, Marchand, and Lempitsky}]{ganin2016domain}
Ganin, Y.; Ustinova, E.; Ajakan, H.; Germain, P.; Larochelle, H.; Laviolette, F.; Marchand, M.; and Lempitsky, V. 2016.
\newblock Domain-adversarial training of neural networks.
\newblock \emph{The journal of machine learning research}, 17(1): 2096--2030.

\bibitem[{Ge et~al.(2023)Ge, Ren, Xu, and Yan}]{ge2023unsupervised}
Ge, P.; Ren, C.-X.; Xu, X.-L.; and Yan, H. 2023.
\newblock Unsupervised domain adaptation via deep conditional adaptation network.
\newblock \emph{Pattern Recognition}, 134: 109088.

\bibitem[{Gretton et~al.(2012)Gretton, Borgwardt, Rasch, Sch{\"o}lkopf, and Smola}]{gretton2012kernel}
Gretton, A.; Borgwardt, K.~M.; Rasch, M.~J.; Sch{\"o}lkopf, B.; and Smola, A. 2012.
\newblock A kernel two-sample test.
\newblock \emph{The Journal of Machine Learning Research}, 13(1): 723--773.

\bibitem[{Grill et~al.(2020)Grill, Strub, Altch{\'e}, Tallec, Richemond, Buchatskaya, Doersch, Avila~Pires, Guo, Gheshlaghi~Azar et~al.}]{grill2020bootstrap}
Grill, J.-B.; Strub, F.; Altch{\'e}, F.; Tallec, C.; Richemond, P.; Buchatskaya, E.; Doersch, C.; Avila~Pires, B.; Guo, Z.; Gheshlaghi~Azar, M.; et~al. 2020.
\newblock Bootstrap your own latent-a new approach to self-supervised learning.
\newblock \emph{Advances in neural information processing systems}, 33: 21271--21284.

\bibitem[{He, Liu, and Yin(2024)}]{he2024enhancing}
He, J.; Liu, B.; and Yin, G. 2024.
\newblock Enhancing Semi-supervised Domain Adaptation via Effective Target Labeling.
\newblock In \emph{Proceedings of the AAAI Conference on Artificial Intelligence}, volume~38, 12385--12393.

\bibitem[{He et~al.(2020)He, Fan, Wu, Xie, and Girshick}]{he2020momentum}
He, K.; Fan, H.; Wu, Y.; Xie, S.; and Girshick, R. 2020.
\newblock Momentum contrast for unsupervised visual representation learning.
\newblock In \emph{Proceedings of the IEEE/CVF conference on computer vision and pattern recognition}, 9729--9738.

\bibitem[{He et~al.(2016)He, Zhang, Ren, and Sun}]{he2016deep}
He, K.; Zhang, X.; Ren, S.; and Sun, J. 2016.
\newblock Deep residual learning for image recognition.
\newblock In \emph{Proceedings of the IEEE conference on computer vision and pattern recognition}, 770--778.

\bibitem[{Hoffman et~al.(2018)Hoffman, Tzeng, Park, Zhu, Isola, Saenko, Efros, and Darrell}]{hoffman2018cycada}
Hoffman, J.; Tzeng, E.; Park, T.; Zhu, J.-Y.; Isola, P.; Saenko, K.; Efros, A.; and Darrell, T. 2018.
\newblock Cycada: Cycle-consistent adversarial domain adaptation.
\newblock In \emph{International conference on machine learning}, 1989--1998. Pmlr.

\bibitem[{Huang, Zhu, and Chen(2023)}]{huang2023semi}
Huang, X.; Zhu, C.; and Chen, W. 2023.
\newblock Semi-supervised Domain Adaptation via Prototype-based Multi-level Learning.
\newblock \emph{arXiv preprint arXiv:2305.02693}.

\bibitem[{Jiang et~al.(2020)Jiang, Wu, Han, Shao, Qi, and Li}]{jiang2020bidirectional}
Jiang, P.; Wu, A.; Han, Y.; Shao, Y.; Qi, M.; and Li, B. 2020.
\newblock Bidirectional Adversarial Training for Semi-Supervised Domain Adaptation.
\newblock In \emph{IJCAI}, 934--940.

\bibitem[{Kim and Kim(2020)}]{kim2020attract}
Kim, T.; and Kim, C. 2020.
\newblock Attract, perturb, and explore: Learning a feature alignment network for semi-supervised domain adaptation.
\newblock In \emph{European conference on computer vision}, 591--607. Springer.

\bibitem[{Krizhevsky, Sutskever, and Hinton(2012)}]{krizhevsky2012imagenet}
Krizhevsky, A.; Sutskever, I.; and Hinton, G.~E. 2012.
\newblock Imagenet classification with deep convolutional neural networks.
\newblock \emph{Advances in neural information processing systems}, 25.

\bibitem[{Krizhevsky, Sutskever, and Hinton(2017)}]{krizhevsky2017imagenet}
Krizhevsky, A.; Sutskever, I.; and Hinton, G.~E. 2017.
\newblock ImageNet classification with deep convolutional neural networks.
\newblock \emph{Communications of the ACM}, 60(6): 84--90.

\bibitem[{Li et~al.(2021{\natexlab{a}})Li, Li, Shi, and Yu}]{li2021cross}
Li, J.; Li, G.; Shi, Y.; and Yu, Y. 2021{\natexlab{a}}.
\newblock Cross-domain adaptive clustering for semi-supervised domain adaptation.
\newblock In \emph{Proceedings of the IEEE/CVF Conference on Computer Vision and Pattern Recognition}, 2505--2514.

\bibitem[{Li, Li, and Yu(2023)}]{li2023adaptive}
Li, J.; Li, G.; and Yu, Y. 2023.
\newblock Adaptive betweenness clustering for semi-supervised domain adaptation.
\newblock \emph{IEEE Transactions on Image Processing}.

\bibitem[{Li, Li, and Yu(2024)}]{li2024inter}
Li, J.; Li, G.; and Yu, Y. 2024.
\newblock Inter-domain mixup for semi-supervised domain adaptation.
\newblock \emph{Pattern Recognition}, 146: 110023.

\bibitem[{Li et~al.(2021{\natexlab{b}})Li, Liu, Zhao, Zhang, and Fu}]{li2021ecacl}
Li, K.; Liu, C.; Zhao, H.; Zhang, Y.; and Fu, Y. 2021{\natexlab{b}}.
\newblock ECACL: A holistic framework for semi-supervised domain adaptation.
\newblock In \emph{Proceedings of the IEEE/CVF International Conference on Computer Vision}, 8578--8587.

\bibitem[{Liang, Hu, and Feng(2020)}]{liang2020we}
Liang, J.; Hu, D.; and Feng, J. 2020.
\newblock Do we really need to access the source data? source hypothesis transfer for unsupervised domain adaptation.
\newblock In \emph{International conference on machine learning}, 6028--6039. PMLR.

\bibitem[{Liu et~al.(2020)Liu, Niles-Weed, Razavian, and Fernandez-Granda}]{liu2020early}
Liu, S.; Niles-Weed, J.; Razavian, N.; and Fernandez-Granda, C. 2020.
\newblock Early-learning regularization prevents memorization of noisy labels.
\newblock \emph{Advances in neural information processing systems}, 33: 20331--20342.

\bibitem[{Long et~al.(2018)Long, Cao, Wang, and Jordan}]{long2018conditional}
Long, M.; Cao, Z.; Wang, J.; and Jordan, M.~I. 2018.
\newblock Conditional adversarial domain adaptation.
\newblock \emph{Advances in neural information processing systems}, 31.

\bibitem[{Long et~al.(2017)Long, Zhu, Wang, and Jordan}]{long2017deep}
Long, M.; Zhu, H.; Wang, J.; and Jordan, M.~I. 2017.
\newblock Deep transfer learning with joint adaptation networks.
\newblock In \emph{International conference on machine learning}, 2208--2217. PMLR.

\bibitem[{Ma et~al.(2022)Ma, Bu, Lu, Wen, Zhou, Zhang, Gu, Li, and Yan}]{ma2022context}
Ma, N.; Bu, J.; Lu, L.; Wen, J.; Zhou, S.; Zhang, Z.; Gu, J.; Li, H.; and Yan, X. 2022.
\newblock Context-guided entropy minimization for semi-supervised domain adaptation.
\newblock \emph{Neural Networks}, 154: 270--282.

\bibitem[{Ngo et~al.(2023)Ngo, Chae, Kwon, Park, and Cho}]{ngo2023improved}
Ngo, B.~H.; Chae, Y.~J.; Kwon, J.~E.; Park, J.~H.; and Cho, S.~I. 2023.
\newblock Improved Knowledge Transfer for Semi-supervised Domain Adaptation via Trico Training Strategy.
\newblock In \emph{Proceedings of the IEEE/CVF International Conference on Computer Vision}, 19214--19223.

\bibitem[{Oord, Li, and Vinyals(2018)}]{oord2018representation}
Oord, A. v.~d.; Li, Y.; and Vinyals, O. 2018.
\newblock Representation learning with contrastive predictive coding.
\newblock \emph{arXiv preprint arXiv:1807.03748}.

\bibitem[{Pan et~al.(2010)Pan, Tsang, Kwok, and Yang}]{pan2010domain}
Pan, S.~J.; Tsang, I.~W.; Kwok, J.~T.; and Yang, Q. 2010.
\newblock Domain adaptation via transfer component analysis.
\newblock \emph{IEEE transactions on neural networks}, 22(2): 199--210.

\bibitem[{Peng et~al.(2019)Peng, Bai, Xia, Huang, Saenko, and Wang}]{peng2019moment}
Peng, X.; Bai, Q.; Xia, X.; Huang, Z.; Saenko, K.; and Wang, B. 2019.
\newblock Moment matching for multi-source domain adaptation.
\newblock In \emph{Proceedings of the IEEE International Conference on Computer Vision}, 1406--1415.

\bibitem[{Qin et~al.(2021)Qin, Wang, Ma, Yin, Wang, and Fu}]{qin2021contradictory}
Qin, C.; Wang, L.; Ma, Q.; Yin, Y.; Wang, H.; and Fu, Y. 2021.
\newblock Contradictory structure learning for semi-supervised domain adaptation.
\newblock In \emph{Proceedings of the 2021 SIAM International Conference on Data Mining (SDM)}, 576--584. SIAM.

\bibitem[{Qin et~al.(2022)Qin, Wang, Ma, Yin, Wang, and Fu}]{qin2022semi}
Qin, C.; Wang, L.; Ma, Q.; Yin, Y.; Wang, H.; and Fu, Y. 2022.
\newblock Semi-Supervised Domain Adaptive Structure Learning.
\newblock \emph{IEEE Transactions on Image Processing}, 31: 7179--7190.

\bibitem[{Rawat and Wang(2017)}]{rawat2017deep}
Rawat, W.; and Wang, Z. 2017.
\newblock Deep convolutional neural networks for image classification: A comprehensive review.
\newblock \emph{Neural computation}, 29(9): 2352--2449.

\bibitem[{Saenko et~al.(2010)Saenko, Kulis, Fritz, and Darrell}]{saenko2010adapting}
Saenko, K.; Kulis, B.; Fritz, M.; and Darrell, T. 2010.
\newblock Adapting visual category models to new domains.
\newblock In \emph{Computer Vision--ECCV 2010: 11th European Conference on Computer Vision, Heraklion, Crete, Greece, September 5-11, 2010, Proceedings, Part IV 11}, 213--226. Springer.

\bibitem[{Saito et~al.(2019)Saito, Kim, Sclaroff, Darrell, and Saenko}]{saito2019semi}
Saito, K.; Kim, D.; Sclaroff, S.; Darrell, T.; and Saenko, K. 2019.
\newblock Semi-supervised domain adaptation via minimax entropy.
\newblock In \emph{Proceedings of the IEEE/CVF International Conference on Computer Vision}, 8050--8058.

\bibitem[{Shen et~al.(2018)Shen, Qu, Zhang, and Yu}]{shen2018wasserstein}
Shen, J.; Qu, Y.; Zhang, W.; and Yu, Y. 2018.
\newblock Wasserstein distance guided representation learning for domain adaptation.
\newblock In \emph{Proceedings of the AAAI Conference on Artificial Intelligence}, volume~32.

\bibitem[{Simonyan and Zisserman(2014)}]{simonyan2014very}
Simonyan, K.; and Zisserman, A. 2014.
\newblock Very deep convolutional networks for large-scale image recognition.
\newblock \emph{arXiv preprint arXiv:1409.1556}.

\bibitem[{Singh(2021)}]{singh2021clda}
Singh, A. 2021.
\newblock CLDA: Contrastive Learning for Semi-Supervised Domain Adaptation.
\newblock In Ranzato, M.; Beygelzimer, A.; Dauphin, Y.; Liang, P.; and Vaughan, J.~W., eds., \emph{Advances in Neural Information Processing Systems}, volume~34, 5089--5101. Curran Associates, Inc.

\bibitem[{Singh et~al.(2021)Singh, Doraiswamy, Takamuku, Bhalerao, Dutta, Biswas, Chepuri, Vengatesan, and Natori}]{singh2021improving}
Singh, A.; Doraiswamy, N.; Takamuku, S.; Bhalerao, M.; Dutta, T.; Biswas, S.; Chepuri, A.; Vengatesan, B.; and Natori, N. 2021.
\newblock Improving semi-supervised domain adaptation using effective target selection and semantics.
\newblock In \emph{Proceedings of the IEEE/CVF Conference on Computer Vision and Pattern Recognition}, 2709--2718.

\bibitem[{Song et~al.(2019)Song, Kim, Park, and Lee}]{song2019prestopping}
Song, H.; Kim, M.; Park, D.; and Lee, J.-G. 2019.
\newblock Prestopping: How does early stopping help generalization against label noise?

\bibitem[{Sun et~al.(2022)Sun, Lu, Zhang, and Ling}]{sun2022safe}
Sun, T.; Lu, C.; Zhang, T.; and Ling, H. 2022.
\newblock Safe self-refinement for transformer-based domain adaptation.
\newblock In \emph{Proceedings of the IEEE/CVF conference on computer vision and pattern recognition}, 7191--7200.

\bibitem[{Touvron et~al.(2021)Touvron, Cord, Douze, Massa, Sablayrolles, and J{\'e}gou}]{touvron2021training}
Touvron, H.; Cord, M.; Douze, M.; Massa, F.; Sablayrolles, A.; and J{\'e}gou, H. 2021.
\newblock Training data-efficient image transformers \& distillation through attention.
\newblock In \emph{International conference on machine learning}, 10347--10357. PMLR.

\bibitem[{Vaswani et~al.(2017)Vaswani, Shazeer, Parmar, Uszkoreit, Jones, Gomez, Kaiser, and Polosukhin}]{vaswani2017attention}
Vaswani, A.; Shazeer, N.; Parmar, N.; Uszkoreit, J.; Jones, L.; Gomez, A.~N.; Kaiser, {\L}.; and Polosukhin, I. 2017.
\newblock Attention is all you need.
\newblock \emph{Advances in neural information processing systems}, 30.

\bibitem[{Venkateswara et~al.(2017)Venkateswara, Eusebio, Chakraborty, and Panchanathan}]{venkateswara2017deep}
Venkateswara, H.; Eusebio, J.; Chakraborty, S.; and Panchanathan, S. 2017.
\newblock Deep hashing network for unsupervised domain adaptation.
\newblock In \emph{Proceedings of the IEEE Conference on Computer Vision and Pattern Recognition}, 5018--5027.

\bibitem[{Wang et~al.(2022)Wang, Peng, Yang, Yang, Zhu, Wang, and You}]{wang2022reliable}
Wang, K.; Peng, X.; Yang, S.; Yang, J.; Zhu, Z.; Wang, X.; and You, Y. 2022.
\newblock Reliable label correction is a good booster when learning with extremely noisy labels.
\newblock \emph{arXiv preprint arXiv:2205.00186}.

\bibitem[{Wang and Qi(2022)}]{wang2022contrastive}
Wang, X.; and Qi, G.-J. 2022.
\newblock Contrastive learning with stronger augmentations.
\newblock \emph{IEEE transactions on pattern analysis and machine intelligence}, 45(5): 5549--5560.

\bibitem[{Xie et~al.(2018)Xie, Zheng, Chen, and Chen}]{xie2018learning}
Xie, S.; Zheng, Z.; Chen, L.; and Chen, C. 2018.
\newblock Learning semantic representations for unsupervised domain adaptation.
\newblock In \emph{International conference on machine learning}, 5423--5432. PMLR.

\bibitem[{Xu et~al.(2021)Xu, Chen, Wang, Wang, Li, and Jin}]{xu2021cdtrans}
Xu, T.; Chen, W.; Wang, P.; Wang, F.; Li, H.; and Jin, R. 2021.
\newblock Cdtrans: Cross-domain transformer for unsupervised domain adaptation.
\newblock \emph{arXiv preprint arXiv:2109.06165}.

\bibitem[{Yan et~al.(2022)Yan, Wu, Li, Qin, Han, and Cui}]{yan2022multi}
Yan, Z.; Wu, Y.; Li, G.; Qin, Y.; Han, X.; and Cui, S. 2022.
\newblock Multi-level consistency learning for semi-supervised domain adaptation.
\newblock \emph{arXiv preprint arXiv:2205.04066}.

\bibitem[{Yang et~al.(2023)Yang, Liu, Xu, and Huang}]{yang2023tvt}
Yang, J.; Liu, J.; Xu, N.; and Huang, J. 2023.
\newblock Tvt: Transferable vision transformer for unsupervised domain adaptation.
\newblock In \emph{Proceedings of the IEEE/CVF Winter Conference on Applications of Computer Vision}, 520--530.

\bibitem[{Yang et~al.(2021)Yang, Wang, Gao, Shrivastava, Weinberger, Chao, and Lim}]{yang2021deep}
Yang, L.; Wang, Y.; Gao, M.; Shrivastava, A.; Weinberger, K.~Q.; Chao, W.-L.; and Lim, S.-N. 2021.
\newblock Deep co-training with task decomposition for semi-supervised domain adaptation.
\newblock In \emph{Proceedings of the IEEE/CVF International Conference on Computer Vision}, 8906--8916.

\bibitem[{Yao et~al.(2018)Yao, Wang, Tsang, Zhang, Sun, Zhang, and Zhang}]{yao2018deep}
Yao, J.; Wang, J.; Tsang, I.~W.; Zhang, Y.; Sun, J.; Zhang, C.; and Zhang, R. 2018.
\newblock Deep learning from noisy image labels with quality embedding.
\newblock \emph{IEEE Transactions on Image Processing}, 28(4): 1909--1922.

\bibitem[{Yu and Lin(2023)}]{yu2023semi}
Yu, Y.-C.; and Lin, H.-T. 2023.
\newblock Semi-Supervised Domain Adaptation with Source Label Adaptation.
\newblock In \emph{Proceedings of the IEEE/CVF Conference on Computer Vision and Pattern Recognition}, 24100--24109.

\bibitem[{Zhang et~al.(2017)Zhang, Cisse, Dauphin, and Lopez-Paz}]{zhang2017mixup}
Zhang, H.; Cisse, M.; Dauphin, Y.~N.; and Lopez-Paz, D. 2017.
\newblock mixup: Beyond empirical risk minimization.
\newblock \emph{arXiv preprint arXiv:1710.09412}.

\bibitem[{Zhang and Ma(2022)}]{zhang2022rethinking}
Zhang, J.; and Ma, K. 2022.
\newblock Rethinking the augmentation module in contrastive learning: Learning hierarchical augmentation invariance with expanded views.
\newblock In \emph{Proceedings of the IEEE/CVF Conference on Computer Vision and Pattern Recognition}, 16650--16659.

\bibitem[{Zhang et~al.(2023)Zhang, Wang, Li, Zhuang, and Lin}]{zhang2023towards}
Zhang, Y.; Wang, Z.; Li, J.; Zhuang, J.; and Lin, Z. 2023.
\newblock Towards Effective Instance Discrimination Contrastive Loss for Unsupervised Domain Adaptation.
\newblock In \emph{Proceedings of the IEEE/CVF International Conference on Computer Vision}, 11388--11399.

\bibitem[{Zhu, Bai, and Wang(2023)}]{zhu2023patch}
Zhu, J.; Bai, H.; and Wang, L. 2023.
\newblock Patch-Mix Transformer for Unsupervised Domain Adaptation: A Game Perspective.
\newblock In \emph{Proceedings of the IEEE/CVF Conference on Computer Vision and Pattern Recognition}, 3561--3571.

\end{thebibliography}

\end{document}